\pdfoutput=1
\PassOptionsToPackage{table}{xcolor}

\documentclass[11pt]{article}

\usepackage{acl}
\usepackage{times}
\usepackage{latexsym}
\usepackage{graphicx}
\usepackage{tabularx}

\usepackage{caption}
\usepackage{subcaption}
\usepackage{amsmath}
\usepackage{arydshln}

\usepackage[linguistics]{forest} 
\usepackage{algorithmic,algorithm}
\renewcommand{\algorithmiccomment}[1]{\bgroup\hfill$\triangleright$~#1\egroup}

\usepackage{arydshln}

\usepackage[T1]{fontenc}

\usepackage[utf8]{inputenc}

\usepackage{microtype}

\usepackage{inconsolata}

\usepackage[table]{xcolor}

\usepackage{synttree}

\title{\texttt{jp-evalb}: Robust Alignment-based PARSEVAL Measures}

\author{
{Jungyeul Park}$^{1}$~~~~
Junrui Wang$^{1}$~~~~
Eunkyul Leah Jo$^{2,3}$~~~~ 
Angela Yoonseo Park$^{1}$\\
$^{1}$Department of Linguistics, The University of British Columbia, Canada\\
$^{2}$Department of Computer Science, The University of British Columbia, Canada\\
$^{3}$Faculté des Sciences et Ingénierie, Sorbonne Université, France\\
{\tt jungyeul@mail.ubc.ca}~~~ {\tt \{wjr0710,eunkyul,apark03\}@student.ubc.ca}\\
}

\begin{document}
\maketitle
\begin{abstract}

We introduce an evaluation system designed to compute PARSEVAL measures, offering a viable alternative to \texttt{evalb} commonly used for constituency parsing evaluation.
The widely used \texttt{evalb} script has traditionally been employed for evaluating the accuracy of constituency parsing results, albeit with the requirement for consistent tokenization and sentence boundaries. 
In contrast, our approach, named \texttt{jp-evalb}, is founded on an alignment method. This method aligns sentences and words when discrepancies arise.
It aims to overcome several known issues associated with \texttt{evalb} by utilizing the `jointly preprocessed (JP)' alignment-based method.
We introduce a more flexible and adaptive framework, ultimately contributing to a more accurate assessment of constituency parsing performance. 
\end{abstract}

\section{Introduction}

For constituency parsing, whether statistical or neural, we rely on the \texttt{evalb} implementation\footnote{\url{http://nlp.cs.nyu.edu/evalb}}, which implements the PARSEVAL measures \citep{black-etal-1991-procedure} as the standard method for evaluating parser performance. 
There is also a variant of the \texttt{evalb\_spmrl} implementation specifically designed for the SPMRL shared task, allowing the evaluation to consider functional phrase labels \citep{seddah-EtAl:2013:SPMRL,seddah-kubler-tsarfaty:2014}. 
A constituent in a hypothesis parse of a sentence is labeled as correct if it matches a constituent in the reference parse with the same non-terminal symbol and span (starting and end indexes). Despite its success in evaluating language technology, \texttt{evalb} faces unresolved critical issues in our discipline.
\texttt{evalb} imposes constraints, demanding consistent tokenization and sentence boundary outcomes. Its implementation assumes equal-length gold and system files with matching terminal nodes. 

In machine translation (MT), sentence alignment involves identifying corresponding sentences in two or more languages and linking sentences from one language to their corresponding counterparts in another. Sentence alignment has been a subject of study for many years, leading to the development of various algorithms.
Early research in this area relied on statistical methods that used bilingual corpora to create models capturing the lexical equivalence between words in different languages. For instance, the Gale-Church algorithm, based on sentence length, was one such approach \citep{gale-church:1993:CL}. \texttt{Bleualign} introduced a more advanced iterative bootstrapping approach building on length-based methods \citep{sennrich-volk-2011-iterative}.
Earlier approaches also aimed to enhance sentence alignment methodologies by incorporating lexical correspondences, as seen in \texttt{hunalign} \citep{varga-EtAl:2005} or the IBM-model based lexicon translation approach \citep{moore:2002}. 
Some attempts involved the integration of linguistic knowledge, heuristics, and various scoring methods to improve efficiency, as demonstrated by \texttt{vecalign} \citep{thompson-koehn-2019-vecalign}.
Word alignment methodologies are also employed to establish correspondences between words in one language and their direct translations in another. Widely used IBM models \citep{brown-EtAl:1993:CL}, along with tools like \texttt{giza++} \citep{och-ney:2000:ACL,och-ney:2003:CL} or \texttt{BerkeleyAligner} \citep{liang-etal-2006-alignment,denero-klein-2007-tailoring}, are capable of aligning words.

Syntactic analysis in the current field of language technology has been predominantly reliant on dependencies. 
Semantic parsing in its higher-level analyses often relies heavily on dependency structures as well.
Therefore, dependency parsing and its evaluation method have their own advantages, such as a more direct representation of grammatical relations and often simpler parsing algorithms.
However, constituency parsing maintains the hierarchical structure of a sentence, which can still be valuable for understanding the syntactic relationships between words and phrases.
Various studies on formal syntax have focused on constituent structures, such as combinatory categorial grammar (CCG) parsing \citep{lewis-lee-zettlemoyer:2016:NAACL,lee-lewis-zettlemoyer:2016:EMNLP,stanojevic-steedman:2020:ACL,yamaki-etal-2023-holographic} or tree-adjoining grammar (TAG) parsing \citep{kasai-etal-2017-tag,kasai-etal-2018-end} (whereas CCG and TAG also inherently incorporate dependency structures). In addition, there have been ongoing studies on constituency parsing, such as the linearization parsing method \citep{vinyals-EtAl:2015:NIPS,liu-zhang-2017-encoder,liu-zhang-2017-shift,fernandez-gonzalez-gomez-rodriguez-2020-enriched,wei-etal-2020-span}.
If a method that utilizes constituent structures is designed to achieve the goal of creating an end-to-end system, it requires more robust evaluation methods for their constituent structure evaluation.

This paper builds upon our recently introduced alignment-based algorithm, for computing PARSEVAL measures \citep{jo-park-park-2024-evalb}, which offers a novel approach for calculating precision, recall, and F scores, even in cases of sentence and word mismatch. 
The primary objective of this paper is to replicate the outcomes generated by \texttt{evalb} during the evaluation process. This aims to achieve a comprehensive understanding of the parser's performance by addressing the previous issues of \texttt{evalb} and preserving its long-standing legacy. 
It includes the numbers of gold, test, matched brackets, and cross brackets, as well as precision, recall, and F scores. Furthermore, we present the number of correct POS tags and their tagging accuracy, following a methodology employed by \texttt{evalb}.
Our proposed method \texttt{jp-evalb} is particularly crucial in end-to-end settings, where deviations from the gold file may arise due to variations in tokenization and sentence boundary results.

\section{Detailing the \texttt{jp-evalb} Algorithm}

To describe the proposed algorithms, we use the following notations for conciseness and simplicity. $\mathcal{T_{L}}$ and $\mathcal{T_{R}}$ introduce the entire parse trees of gold and system files, respectively. 
$\mathcal{T_{L}}$ is a simplified notation representing $\mathcal{T}_{\mathcal{L}(l)}$, where $l$ is the list of tokens in $\mathcal{L}$. This notation applies in the same manner to $\mathcal{R}$.
$\mathcal{S}_{\mathcal{T}}$ represents a set of constituents of a tree $\mathcal{T}$, and $\mathcal{C}(\mathcal{T})$ is the total number constituents of $\mathcal{T}$.
$\mathcal{C}(\texttt{tp})$ is the number of true positive constituents where $\mathcal{S}_{\mathcal{T_{L}}}\cap\mathcal{S}_{\mathcal{T_{R}}}$, and we count it per aligned sentence. 
The presented Algorithm~\ref{alg:main-algorithm} demonstrates the pseudo-code for the new PARSEVAL measures.

\begin{algorithm} [!ht]
\caption{Pseudo-code for \texttt{jp-evalb}}\label{alg:main-algorithm}
{\footnotesize
\begin{algorithmic}[1]
\STATE{\textbf{function} {\textsc{PARSEVALmeasures}} ($\mathcal{T_{L}}$ and $\mathcal{T_{R}}$):}
\begin{ALC@g}
\STATE {Extract the list of tokens $\mathcal{L}$ and $\mathcal{R}$ from $\mathcal{T_{L}}$ and $\mathcal{T_{R}}$}
\STATE {$\mathcal{L^{\prime}}$, $\mathcal{R^{\prime}}$ $\gets$ \textsc{SentenceAlignment}($\mathcal{L}$, $\mathcal{R}$)  
}
\STATE {Align trees based on $\mathcal{L^{\prime}}$ and $\mathcal{R^{\prime}}$ to obtain $\mathcal{T_{L^{\prime}}}$ and $\mathcal{T_{R^{\prime}}}$}

\WHILE{$\mathcal{T}_{\mathcal{L}^{\prime}}$ and $\mathcal{T}_{\mathcal{R}^{\prime}}$}
\STATE {Extract the list of tokens ${l}$ and ${r}$ from $\mathcal{T}_{\mathcal{L}^{\prime}_{i}}$ and $\mathcal{T}_{\mathcal{R}^{\prime}_{i}}$}
\STATE {$l^{\prime}, r^{\prime} \gets {\textsc{WordAlignment}}(l,r)$}
\STATE {$\mathcal{S}_{\mathcal{T}_{\mathcal{L}}}$ $\gets$ \textsc{GetConstituent}($\mathcal{T}_{\mathcal{L}^{\prime}_{i}(l^{\prime})},0$) 
}
\STATE {$\mathcal{S}_{\mathcal{T}_{\mathcal{R}}}$ $\gets$ \textsc{GetConstituent}($\mathcal{T}_{\mathcal{R}^{\prime}_{i}(r^{\prime})},0$) 
}
\STATE {$\mathcal{C}(\mathcal{T_{L}}) \gets \mathcal{C}(\mathcal{T_{L}}) +$ \textsc{len}($\mathcal{S}_{\mathcal{T}_{\mathcal{L}}}$)}
\STATE {$\mathcal{C}(\mathcal{T_{R}}) \gets \mathcal{C}(\mathcal{T_{R}}) +$ \textsc{len}($\mathcal{S}_{\mathcal{T}_{\mathcal{R}}}$)}
\WHILE{$\mathcal{S}_{\mathcal{T}_{\mathcal{L}}}$ {and} $\mathcal{S}_{\mathcal{T}_{\mathcal{R}}}$}
\IF{(\textsc{label}, {\textsc{start}$_{\mathcal{L}}$}, {\textsc{end}$_{\mathcal{L}}$},$l^{\prime}_{j}$)\\
~~~~$=$ (\textsc{label}, {\textsc{start}$_{\mathcal{R}}$}, {\textsc{end}$_{\mathcal{R}}$},$r^{\prime}_{j}$)}
\STATE{$\mathcal{C}(\texttt{tp}) \gets \mathcal{C}(\texttt{tp}) + 1 $  
}
\ENDIF
\ENDWHILE
\ENDWHILE
\RETURN {$\mathcal{C}(\mathcal{T_{L}})$, $\mathcal{C}(\mathcal{T_{R}})$, and $\mathcal{C}(\texttt{tp})$} 
\end{ALC@g}
\end{algorithmic} 
}
\end{algorithm}

\begin{algorithm}[!ht]
\caption{Pseudo-code for alignment}\label{alignment-algorithm}
{\footnotesize
\begin{algorithmic}[1]
\STATE{\textbf{function} \textsc{Alignment} ($\mathcal{L}$, $\mathcal{R}$):}
\begin{ALC@g}
\WHILE{$\mathcal{L}$ and $\mathcal{R}$}
\IF{ Matched {\textsc{cases $_{(i,j)}$}} } 
\STATE{$\mathcal{L^{\prime}}$, $\mathcal{R^{\prime}}$ $\gets$ $\mathcal{L^{\prime}}+\mathcal{L}_i, \mathcal{R^{\prime}}+\mathcal{R}_j$ }
\ELSE
\WHILE{$\neg$(Matched \textsc{cases $_{(i+1,j+1)}$}}
\IF{$\textsc{len}(\mathcal{L}_{i}) < \textsc{len}(\mathcal{R}_{j})$}
\STATE{${L}^{\prime}$ $\gets$ ${L}^{\prime}+\mathcal{L}_{i}$} 
\STATE{$i \gets i+1 $}
\ELSE
\STATE{${R}^{\prime}$ $\gets$ ${R}^{\prime}+\mathcal{R}_{j}$}
\STATE{$j \gets j+1 $}
\ENDIF

\ENDWHILE
\STATE{
$\mathcal{L^{\prime}}$, $\mathcal{R^{\prime}}$ 
$\gets$ 
$\mathcal{L^{\prime}}+L^{\prime}, \mathcal{R^{\prime}}+R^{\prime}$
}
\ENDIF
\ENDWHILE
\RETURN {$\mathcal{L^{\prime}}$, $\mathcal{R^{\prime}}$}
\end{ALC@g}
\end{algorithmic} 
}
\end{algorithm}

In the first stage, we extract leaves $\mathcal{L}$ and $\mathcal{R}$ from the parse trees and align sentences to obtain $\mathcal{L^{\prime}}$ and $\mathcal{R^{\prime}}$ using the sentence alignment algorithm. 
Algorithm~\ref{alignment-algorithm} shows the generic pattern-matching approach of the alignment algorithm where sentence and word alignment can be applied. 
We define the following two cases for matched \textsc{cases$_{(i,j)}$} of sentence alignment:

{
\begin{align}
\mathcal{L}_{i({\not\sqcup})} = \mathcal{R}_{j({\not\sqcup})} \\
\label{sa-case2} (\mathcal{L}_{i({\not\sqcup})} \simeq \mathcal{R}_{j({\not\sqcup})}) ~\land \notag \\
(\mathcal{L}_{i+1({\not\sqcup})} = \mathcal{R}_{j+1({\not\sqcup})} \lor \mathcal{L}_{i+1({\not\sqcup})} \simeq \mathcal{R}_{j+1({\not\sqcup})} )
\end{align}}\noindent 
where we examine whether $\mathcal{L}_{i}$ is similar to or equal ($\simeq$) to $\mathcal{R}_{j}$ based on the condition that the ratio of edit distance to the entire character length is less than 0.1 in ~\eqref{sa-case2}.
While the necessity of sentence alignment is rooted in a common phenomenon in cross-language tasks {such as machine translation}, the intralingual alignment between gold and system sentences does not share the same necessity {because $\mathcal{L}$ and $\mathcal{R}$ are} identical sentences that only differ in sentence boundaries and token.
A notation ${\not\sqcup}$ is introduced to represent spaces that are removed {during sentence alignment when comparing $\mathcal{L}{i}$ and $\mathcal{R}_{j}$, irrespective of their tokenization results.}
If there is a mismatch due to differences in sentence boundaries, the algorithm accumulates the sentences until the next pair of sentences represented as \textsc{case $n$ ${(i+1,j+1)}$}, is matched.

In the next stage {of Algorithm~\ref{alg:main-algorithm},} 
we align trees based on $\mathcal{L}^{\prime}$ and $\mathcal{R}^{\prime}$ to obtain $\mathcal{T}_{\mathcal{L}^{\prime}}$ and $\mathcal{T}_{\mathcal{R}^{\prime}}$. 
By iterating through $\mathcal{T}_{\mathcal{L}^{\prime}}$ and $\mathcal{T}_{\mathcal{R}^{\prime}}$, we conduct word alignment and compare pairs of sets of constituents for each corresponding pair of $\mathcal{T}_{\mathcal{L}^{\prime}_{i}}$ and {$\mathcal{T}_{\mathcal{R}^{\prime}_{j}}$}.
The word alignment algorithm adopts a logic similar to sentence alignment. 
It involves the accumulation of words in $l^{\prime}$ and $r^{\prime}$ under the condition that pairs of $l_i$ and $r_j$ do not match, often attributed to tokenization mismatches. Here, we assume interchangeability between notations of sentence alignment ($\mathcal{L}$) and word alignment ($l_i$).
We define the following two cases for matched \textsc{cases$_{(i,j)}$} of word alignment:
{
\begin{align}
\label{wa-case1} l_{i} = r_{j} \\
\label{wa-case2} (l_{i} \neq r_{j}) \land (l_{i+1} = r_{j+1}) 
\end{align}}

When deciding whether to accumulate the token from $l_{i+1}$ or $r_{j+1}$ in the case of a word mismatch, we base our decision on the following condition, rather than a straightforward comparison between the lengths of the current tokens $l_{i}$ and $r_{j}$:
$(\textsc{len}(l) - \textsc{len}(l_{0..i}))  > (\textsc{len}(r) - \textsc{len}(r_{0..j}))$




Finally, we extract a set of constituents, a straightforward procedure for obtaining constituents from a given tree, which includes the label name, start index, end index, and a list of tokens.
The current proposed method utilizes simple pattern matching for sentence and word alignment, operating under the assumption that the gold and system sentences are the same, with minimal potential for morphological mismatches.
This differs from sentence and word alignment in machine translation.
MT usually relies on recursive editing and EM algorithms due to the inherent difference between source and target languages.

\section{Word and Sentence Mismatches}

\paragraph{Word mismatch}
We have observed that the expression of contractions varies significantly, resulting in inherent challenges related to word mismatches.
As the number of contractions and symbols to be converted in a language is finite, we composed an exception list for our system to capture such cases for each language to facilitate the word alignment process between gold and system sentences.
In the following example, we achieve perfect precision and recall of 5/5 for both because their constituent trees are exactly matched, regardless of any mismatched words.
If the word mismatch example is not in the exception list, we perform the word alignment. We can still achieve perfect precision and recall (5/5 for both) without the word mismatch exception list because their constituent trees can be exactly matched based on the word-alignment of \{$^{1.0}$\textit{ca}  $^{1.1}$\textit{n't}\} and \{$^{1.0}$\textit{can}  $^{1.1}$\textit{not}\} (Figure~\ref{example-word-mismatch}).

\begin{center}
{\footnotesize
\begin{tabular}{r c cc c c }
gold &  $^{0}$\textit{This} & $^{1.0}$\textit{ca} & $^{1.1}$\textit{n't} & $^{2}$\textit{be} & $^{3}$\textit{right}\\
system & $^{0}$\textit{this} & $^{1.0}$\textit{can} & $^{1.1}$\textit{not} & $^{2}$\textit{be} & $^{3}$\textit{right}\\ 
\end{tabular}
}
\end{center}

The effectiveness of the word alignment approach remains intact even for morphological mismatches where "morphological segmentation is not the inverse of concatenation" \citep{tsarfaty-nivre-andersson:2012:ACL2012short}, such as in morphologically rich languages.  
For example, we trace back to the sentence in Hebrew described in \citet{tsarfaty-nivre-andersson:2012:ACL2012short} as a word mismatch example caused by morphological analyses:

\begin{center}
\resizebox{.49\textwidth}{!}{
{\footnotesize
\begin{tabular}{r ccc cccc}
gold & $^{0}$\textit{B} & $^{1.0}$\textit{H} & $^{1.1}$\textit{CL} & $^{2}$\textit{FL} & $^{3}$\textit{HM} & $^{4.0}$\textit{H} & $^{4.1}$\textit{NEIM} \\
& 'in' & 'the' & 'shadow' & 'of' & 'them' & 'the' & 'pleasant' \\
system & $^{0}$\textit{B} & \multicolumn{2}{c}{$^{1}$\textit{CL}} & $^{2}$\textit{FL} & $^{3}$\textit{HM} & \multicolumn{2}{c}{$^{4}$\textit{HNEIM}} \\ 
& 'in' & \multicolumn{2}{c}{'shadow'} & 'of' & 'them' & \multicolumn{2}{c}{'made-pleasant'} \\
\end{tabular}
}}
\end{center}
Pairs of \{$^{1.0}$\textit{H} $^{1.1}$\textit{CL}, $^{1}$\textit{CL}\} ('the shadow') and \{$^{4.0}$\textit{H} $^{4.1}$\textit{NEIM}, $^{4}$\textit{HNEIM}\} ('the pleasant') are word-aligned using the proposed algorithm, resulting in a precision of 4/4 and recall of 4/6 (Figure~\ref{example-word-mismatch-additional}).

\paragraph{Sentence mismatch}
When there are sentence mismatches, they would be aligned and merged as a single tree using a dummy root node: for example, \textsc{@s} which can be ignored during evaluation.
In the following example, we obtain precision of 5/8 and recall of 5/7 (Figure~\ref{example-sentence-mismatch}).

\begin{figure*} [!ht]
  \centering
  \begin{subfigure}[b]{\textwidth}
 \centering
\begin{center}
\resizebox{\textwidth}{!}{\footnotesize
\begin{tabular}{p{1cm}p{1cm}p{1cm}p{1cm}p{1cm}p{1cm} cc | cc p{1cm}p{1cm}p{1cm}p{1cm}p{1cm}p{1cm}}
&& \multicolumn{6}{c}{(gold)} & \multicolumn{6}{c}{(system)} &&\\

&&\multicolumn{1}{c}{\cellcolor{blue!15}S$_{(0,4)}$}&  & &  \multicolumn{1}{c}{\cellcolor{yellow!15}NP$_{(0,1)}$} &DT & \textit{$^{0}$This}  & \textit{$^{0}$this} & DT & \multicolumn{1}{c}{\cellcolor{yellow!15}NP$_{(0,1)}$}  & & & \multicolumn{1}{c}{\cellcolor{blue!15}S$_{(0,4)}$}&&\\

&&\cellcolor{blue!15} &\multicolumn{1}{c}{\cellcolor{red!15}VP$_{(1,4)}$}& &  &MD & \textit{$^{1.0}$ca} & \textit{$^{1.0}$can}  & MD & & & \multicolumn{1}{c}{\cellcolor{red!15}VP$_{(1,4)}$} & \cellcolor{blue!15}&&\\

&&\cellcolor{blue!15} &\cellcolor{red!15}  & &  &RB & \textit{$^{1.1}$n't} & \textit{$^{1.1}$not}  & RB & & & \cellcolor{red!15}& \cellcolor{blue!15}&&\\

&&\cellcolor{blue!15} &\cellcolor{red!15}  & \multicolumn{1}{c}{\cellcolor{green!15}VP$_{(2,4)}$}&  &VB & \textit{$^{2}$be} & \textit{$^{2}$be} & VB & & \multicolumn{1}{c}{\cellcolor{green!15}VP$_{(2,4)}$} & \cellcolor{red!15}& \cellcolor{blue!15}&&\\

&&\cellcolor{blue!15} &\cellcolor{red!15}  & \cellcolor{green!15}  & \multicolumn{1}{c}{\cellcolor{yellow!15}AdjP$_{(3,4)}$}&JJ & \textit{$^{3}$right} & \textit{$^{3}$right}& JJ & \multicolumn{1}{c}{\cellcolor{yellow!15}AdjP$_{(3,4)}$} & \cellcolor{green!15} & \cellcolor{red!15} & \cellcolor{blue!15}&&\\
\end{tabular}
}
\end{center}
 \caption{Example of word mismatches}
 \label{example-word-mismatch}
  \end{subfigure}
  \hfill

  \begin{subfigure}[b]{\textwidth}
 \centering
\begin{center}
\resizebox{\textwidth}{!}{\footnotesize
\begin{tabular}{p{1cm} p{1cm} p{1cm}p{1cm}p{1cm} cc | cc p{1cm}p{1cm}p{1cm}p{1cm} p{1cm}} 

{}& \multicolumn{6}{c}{(gold)} & \multicolumn{6}{c}{(system)} &{}\\

{}& \multicolumn{1}{c}{\cellcolor{blue!15}PP$_{(0,5)}$} & {} & {} & {} & 
'in' & \textit{$^{0}$B} &  \textit{$^{0}$B} & 'in' &  
{} & {} & {} & \multicolumn{1}{c}{\cellcolor{blue!15}PP$_{(0,5)}$} &{}\\

&\multicolumn{1}{c}{\cellcolor{blue!15}} & \multicolumn{1}{c}{\cellcolor{red!15}NP$_{(1,5)}$} & \multicolumn{1}{c}{\cellcolor{green!15}NP$_{(1,4)}$} & {} & 
'the' & \textit{$^{1.0}$H} &  
\textit{} & {} &  
{} & {} & {} & \multicolumn{1}{c}{\cellcolor{blue!15}}& \\

&\multicolumn{1}{c}{\cellcolor{blue!15}} & \multicolumn{1}{c}{\cellcolor{red!15}} & \multicolumn{1}{c}{\cellcolor{green!15}} & {} & 
'shadow' & \textit{$^{1.1}$CL} &  
\textit{$^{1}$CL} & {'shadow'} &  
{} & \multicolumn{1}{c}{\cellcolor{green!15}NP$_{(1,4)}$} & \multicolumn{1}{c}{\cellcolor{red!15}NP$_{(1,5)}$}  & \multicolumn{1}{c}{\cellcolor{blue!15}}&\\

&\multicolumn{1}{c}{\cellcolor{blue!15}} & \multicolumn{1}{c}{\cellcolor{red!15}} & \multicolumn{1}{c}{\cellcolor{green!15}} & \multicolumn{1}{c}{\cellcolor{red!15}PP$_{(2,4)}$} & 
'of' & \textit{$^{2}$FL} &  
\textit{$^{2}$FL} & {'of'} &  
\multicolumn{1}{c}{\cellcolor{red!15}PP$_{(2,4)}$} & \multicolumn{1}{c}{\cellcolor{green!15}} & \multicolumn{1}{c}{\cellcolor{red!15}}  & \multicolumn{1}{c}{\cellcolor{blue!15}}&\\

&\multicolumn{1}{c}{\cellcolor{blue!15}} & \multicolumn{1}{c}{\cellcolor{red!15}} & \multicolumn{1}{c}{\cellcolor{green!15}} & \multicolumn{1}{c}{\cellcolor{red!15}} & 
'them' & \textit{$^{3}$HM} &  
\textit{$^{3}$HM} & {'them'} &  
\multicolumn{1}{c}{\cellcolor{red!15}} & \multicolumn{1}{c}{\cellcolor{green!15}} & \multicolumn{1}{c}{\cellcolor{red!15}}  & \multicolumn{1}{c}{\cellcolor{blue!15}}&\\

&\multicolumn{1}{c}{\cellcolor{blue!15}} & \multicolumn{1}{c}{\cellcolor{red!15}} & {} & \multicolumn{1}{c}{\cellcolor{blue!15}AdjP$_{(4,5)}$} & 
'the' & \textit{$^{4.0}$H} &  
\textit{} & {} &  {} & 
{} & \multicolumn{1}{c}{\cellcolor{red!15}}  & \multicolumn{1}{c}{\cellcolor{blue!15}}&\\

&\multicolumn{1}{c}{\cellcolor{blue!15}} & \multicolumn{1}{c}{\cellcolor{red!15}} & {} & \multicolumn{1}{c}{\cellcolor{blue!15}} & 
'pleasant' & \textit{$^{4.1}$NEIM} &  
\textit{$^{4}$HNEIM} & {'made-pleasant'} &  {} & 
{} & \multicolumn{1}{c}{\cellcolor{red!15}}  & \multicolumn{1}{c}{\cellcolor{blue!15}}&\\

\end{tabular}
}
\end{center}
 \caption{Example of word mismatches with additional morphemes}
 \label{example-word-mismatch-additional}
  \end{subfigure}
  \hfill

  \begin{subfigure}[b]{\textwidth}
 \centering
\begin{center}
\resizebox{\textwidth}{!}{\footnotesize
\begin{tabular}{p{1cm}p{1cm}p{1cm} p{1cm}p{1cm}p{1cm} cc 
| cc p{1cm}p{1cm}p{1cm} p{1cm}p{1cm}p{1cm}  }
\multicolumn{8}{c}{(gold)} & \multicolumn{8}{c}{(system, merged after alignment)} \\

\multicolumn{1}{c}{\cellcolor{blue!15}S$_{(0,6)}$}& \multicolumn{1}{c}{\cellcolor{red!15}S$_{(0,5)}$} & {} &
{} & {} & {} & 
VB & \textit{$^{0}$Click} &  \textit{$^{0}$Click} &  VB & 
{} & \multicolumn{1}{c}{\cellcolor{blue!15}VP$_{(0,2)}$} & \multicolumn{1}{c}{\cellcolor{red!15}S$_{(0,2)}$} & {} & {} & \multicolumn{1}{c}{\cellcolor{blue!15}@S$_{(0,6)}$}\\

\cellcolor{blue!15} & \cellcolor{red!15} & {} & 
{} & {} & \multicolumn{1}{c}{\cellcolor{green!15}AdvP$_{(1,2)}$} & 
RB & \textit{$^{1}$here} &  \textit{$^{1}$here} &  RB & 
\multicolumn{1}{c}{\cellcolor{green!15}AdvP$_{(1,2)}$} & \cellcolor{blue!15} & \cellcolor{red!15} &  & & \cellcolor{blue!15} \\

\cellcolor{blue!15} & \cellcolor{red!15} & \multicolumn{1}{c}{\cellcolor{green!15}S$_{(2,5)}$} & 
\multicolumn{1}{c}{\cellcolor{yellow!15}VP$_{(2,5)}$} & {} & {} & 
TO & \textit{$^{2}$To} &  \textit{$^{2}$To} &  TO & 
&  & \multicolumn{1}{c}{\cellcolor{yellow!15}VP$_{(2,5)}$} & \multicolumn{1}{c}{\cellcolor{green!15}S$_{(2,5)}$} & \multicolumn{1}{c}{\cellcolor{red!15}S$_{(2,6)}$}& \cellcolor{blue!15}\\

\cellcolor{blue!15} & \cellcolor{red!15} & {\cellcolor{green!15}} & 
{\cellcolor{yellow!15}} & \multicolumn{1}{c}{\cellcolor{blue!15}VP$_{(3,5)}$} & {} & 
VB & \textit{$^{3}$view} &  \textit{$^{3}$view} &  VB & 
& \multicolumn{1}{c}{\cellcolor{blue!15}VP$_{(3,5)}$}
& \multicolumn{1}{c}{\cellcolor{yellow!15}} & \multicolumn{1}{c}{\cellcolor{green!15}} & \multicolumn{1}{c}{\cellcolor{red!15}}& \cellcolor{blue!15}\\

\cellcolor{blue!15} & \cellcolor{red!15} & {\cellcolor{green!15}} & 
{\cellcolor{yellow!15}} & \cellcolor{blue!15} & \multicolumn{1}{c}{\cellcolor{green!15}NP$_{(4,5)}$} & 
PRP & \textit{$^{4}$it} &  \textit{$^{4}$it} &  PRP & \multicolumn{1}{c}{\cellcolor{green!15}NP$_{(4,5)}$}
& {\cellcolor{blue!15}}
& {\cellcolor{yellow!15}} & {\cellcolor{green!15}} & {\cellcolor{red!15}}& \cellcolor{blue!15}\\

\cellcolor{blue!15} & & & & &  & .  & $^{5}$. & $^{5}$. & . & 
& & & & \cellcolor{red!15}& \cellcolor{blue!15} \\
\end{tabular}
}
\end{center}

 \caption{Example of sentence mismatches}
 \label{example-sentence-mismatch}
  \end{subfigure}
  \caption{Example of word and sentence mismatches}
  \label{fig:three graphs}
\end{figure*}

\paragraph{Assumptions}
To address morphological analysis discrepancies in the parse tree during evaluation, we establish the following two assumptions: (i) The entire tree constituent can be considered a true positive, even if the morphological segmentation or analysis differs from the gold analysis, as long as the two sentences (gold and system) are aligned and their {root} labels are the same.
(ii) The subtree constituent can be considered a true positive if lexical items align in word alignment, and their phrase labels are the same.

\section{Usage of \texttt{jp-evalb}}

We use the following command to execute the \texttt{jp-evalb} script: 

{\footnotesize
\begin{verbatim}
% python3 jp-evalb.py gold_parsed_file \
                      system_parsed_file
\end{verbatim}
}

\noindent It generates the same output format as \texttt{evalb}.
We provide information for each column in both \texttt{jp-evalb} and \texttt{evalb}, while highlighting their differences. We note that the \texttt{ID}s in \texttt{jp-evalb} may not be exactly the same as in \texttt{evalb} due to the proposed method performing sentence alignment before evaluation.

{
\begin{description} \setlength\itemsep{0em}
\item [\texttt{Sent. ID}, \texttt{Sent. Len.}, \texttt{Stat.}] ID, length, and status of the provided sentence, where status 0 indicates 'OK,' status 1 implies 'skip,' and status 2 represents 'error' for \texttt{evalb}. We do not assign skip or error statuses.
\item [\texttt{Recall}, \texttt{Precision}] Recall and precision of constituents.

\item [\texttt{Matched Bracket}, \texttt{Bracket gold}, \texttt{Bracket test}] Assessment of matched brackets (true positives) in both the gold and test parse trees, and their numbers of constituents. 

\item [\texttt{Cross Bracket}] The number of cross brackets. 

\item [\texttt{Words}, \texttt{Correct Tags}, \texttt{Tag Accuracy}] Evaluation of the number of words, correct POS tags, and POS tagging accuracy. 
\end{description}
}

It's important to note that the original \texttt{evalb} excludes problematic symbols and punctuation marks in the tree structure. Our results include all tokens in the given sentence, and bracket numbers reflect the actual constituents in the system and gold parse trees. 
Accuracy in the last column of the result is determined by comparing the correct number of POS-tagged words to the gold sentence including punctuation marks, differing from the original \texttt{evalb} which doesn't consider word counts or correct POS tags. 
Figure~\ref{comparison} visually depicts the difference in constituent lists between \texttt{jp-evalb} and \texttt{evalb}. 
The original \texttt{evalb} excludes punctuation marks from its consideration of constituents, resulting in our representation of word index numbers in red for \texttt{evalb}. Consequently, \texttt{evalb} displays constituents without punctuation marks and calculates POS tagging accuracy based on six word tokens. On the other hand, \texttt{jp-evalb} includes punctuation marks in constituents and evaluates POS tagging accuracy using eight tokens, which includes two punctuation marks in the sentence. 
We note that the inclusion of punctuation marks in the constituents does not affect the total count, as punctuation marks do not constitute a constituent by themselves. 

Additionally, we offer a legacy option, \texttt{--evalb}, to precisely replicate \texttt{evalb} results. To execute the script with the \texttt{evalb} option, utilize the following command:

{\footnotesize
\begin{verbatim}
% python3 jp-evalb.py gold_parsed_file \
                      system_parsed_file \
                      -evalb param.prm
\end{verbatim}
}

\noindent This option can utilize the default values from the \texttt{COLLINS.prm} file if the parameter file is not provided.
It will accurately reproduce \texttt{evalb} results, even in cases where there are discrepancies such as \texttt{Length unmatch} and \texttt{Words unmatch} errors in \texttt{evalb}'s output. These discrepancies are indicated by the \texttt{Stat.} column, which displays either \texttt{1} (skip) or \texttt{2} (error).

\begin{figure}
    \centering
    
\begin{subfigure}[b]{.4\textwidth}    
\resizebox{\textwidth}{!}
{\footnotesize
\centering
\synttree
[TOP [S [INTJ [RB [$^{0,{\color{red}0}}$No] ]] [, [$^{1}$,]] [NP [PRP [$^{2,{\color{red}1}}$it]]] 
[VP [VBD [$^{3,{\color{red}2}}$was]] [RB [$^{4,{\color{red}3}}$n't]] [NP [NNP [$^{5,{\color{red}4}}$Black]] [NNP [$^{6,{\color{red}5}}$Monday]]]] [$\cdot$ [$^{7}$$\cdot$] ]]]
}
\caption{Example of the parse tree}\label{parse-tree}
\end{subfigure}\hfill

\begin{subfigure}[b]{.5\textwidth}

{\footnotesize
\begin{tabular}{l}
~\\
\texttt{ ('S', 0, 8, "No {,} it was n't Black Monday {.}")}\\
\texttt{ ('INTJ', 0, 1, 'No') }\\
\texttt{ ('NP', 2, 3, 'it') }\\
\texttt{ ('VP', 3, 7, "was n't Black Monday") }\\
\texttt{ ('NP', 5, 7, 'Black Monday') }\\
\end{tabular}
}

\caption{List of constituents by \texttt{jp-evalb}}\label{constituents-jp-evalb}
\end{subfigure}\hfill

\begin{subfigure}[b]{.5\textwidth}
{\footnotesize
\begin{tabular}{l}
~\\
\texttt{('S', {\color{red}0}, {\color{red}6}, "No it was n't Black Monday") }\\
\texttt{('INTJ', {\color{red}0}, {\color{red}1}, "No") }\\
\texttt{('NP', {\color{red}1}, {\color{red}2}, "it")}\\
\texttt{('VP', {\color{red}2}, {\color{red}6}, "was n't Black Monday") }\\
\texttt{('NP', {\color{red}4}, {\color{red}6}, "Black Monday") }\\
\end{tabular}
}

\caption{List of constituents by \texttt{evalb}}\label{constituents-evalb}
\end{subfigure}

\caption{Difference between \texttt{jp-evalb} and \texttt{evalb}}\label{comparison}
\end{figure}

\section{Case Studies}

\paragraph{Section 23 of the English Penn treebank}

Under identical conditions where sentences and words match, the proposed method requires around 4.5 seconds for evaluating the section 23 of the Penn Treebank. 
On the same machine, \texttt{evalb} completes the task less than 0.1 seconds. 
We do not claim that our proposed implementation is fast or faster than \texttt{evalb}, recognizing the well-established differences in performance between compiled languages like C, which \texttt{evalb} used, and interpreted languages such as Python, which our current implementation uses. 
Our proposed method also introduces additional runtime for sentence and word alignment, a process not performed by \texttt{evalb}.
We present excerpts from three result files generated by \texttt{evalb} and our proposed method in Figure~\ref{ptb-section23}. The parsed results were obtained using the PCFG-LA Berkeley parser \citep{petrov-klein:2007:main}. 
It's worth noting that there may be slight variations between the two sets of results because \texttt{evalb} excludes constituents with specific symbols and punctuation marks during evaluation. 
However, as we mentioned earlier, \texttt{jp-evalb} can reproduce the exact same results as \texttt{evalb} for a legacy reason.

\begin{figure}[!ht]
  \centering

\begin{subfigure}[b]{.49\textwidth}
\centering

\resizebox{\textwidth}{!}{\footnotesize
\texttt{\begin{tabular}{rrr rrr rrr rrr}
\multicolumn{12}{c}{}\\
\multicolumn{2}{c}{Sent}& &&& Mt &  \multicolumn{2}{c}{Br} & Cr&& Co& Tag\\
ID&  L&  St& Re&  Pr&  Br& gd& te& Br& Wd&  Tg&  Acc \\
\hdashline
1& 8& 0&  100.00& 100.00&  5&5& 5&0&8&  7& 87.50\\
2&40& 0&70.97&  73.33& 22&  31&30&7&  40& 40&100.00\\
3&31& 0&95.24&  95.24& 20&  21&21&0&  31& 31&100.00\\
4&35& 0&90.48&  86.36& 19&  21&22&2&  35& 35&100.00\\
5&26& 0&86.96&  86.96& 20&  23&23&2&  26& 25& 96.15\\
\multicolumn{12}{l}{.....}
\end{tabular}
}}
\caption{Example of \texttt{jp-evalb} results considering punctuation marks during evaluation}
\label{jp-evalb-results}
\end{subfigure}

\begin{subfigure}[b]{.49\textwidth}
\centering
\resizebox{\textwidth}{!}{\footnotesize
\texttt{\begin{tabular}{rrr rrr rrr rrr}
\multicolumn{2}{c}{Sent}& &&& Mt &  \multicolumn{2}{c}{Br} & Cr&& Co& Tag\\
ID&  L&  St& Re&  Pr&  Br& gd& te& Br& Wd&  Tg&  Acc \\
\hdashline
1& 8& 0&  100.00& 100.00&   5& 5& 5&0&6&  5& 83.33\\
2&40& 0&70.97&  73.33& 22&  31&30&7&  37& 37&100.00\\
3&31& 0&95.24&  95.24& 20&  21&21&0&  26& 26&100.00\\
4&35& 0&90.48&  86.36& 19&  21&22&2&  32& 32&100.00\\
5&26& 0&86.96&  86.96& 20&  23&23&2&  24& 23& 95.83\\
\multicolumn{12}{l}{.....}
\end{tabular}
}}
\caption{Example of \texttt{jp-evalb} results with the legacy option, which produces the exact same results as \texttt{evalb}}
 \label{jp-evalb-legacy-results}
\end{subfigure}

\begin{subfigure}[b]{.49\textwidth}
\centering
\resizebox{\textwidth}{!}{\footnotesize
\texttt{\begin{tabular}{rrr rrr rrr rrr}
\multicolumn{2}{c}{Sent}& &&& Mt &  \multicolumn{2}{c}{Br} & Cr&& Co& Tag\\
ID&  L&  St& Re&  Pr&  Br& gd& te& Br& Wd&  Tg&  Acc \\
\hdashline
1& 8& 0&  100.00& 100.00&   5& 5& 5&0&6&  5& 83.33\\
2&40& 0&70.97&  73.33& 22&  31&30&7&  37& 37&100.00\\
3&31& 0&95.24&  95.24& 20&  21&21&0&  26& 26&100.00\\
4&35& 0&90.48&  86.36& 19&  21&22&2&  32& 32&100.00\\
5&26& 0&86.96&  86.96& 20&  23&23&2&  24& 23& 95.83\\
\multicolumn{12}{l}{.....}
\end{tabular}
}}
 \caption{Example of the original \texttt{evalb} results}
 \label{evalb-results}
\end{subfigure}

\caption{Examples of evaluation results on Section 23 of the English Penn treebank}
\label{ptb-section23}
\end{figure}

\paragraph{Bug cases identified by \texttt{evalb}}

We evaluate bug cases identified by \texttt{evalb}. Figure~\ref{evalb-bug} displays all five identified bug cases, showcasing successful evaluation without any failures. 
In three instances (sentences 1, 2, and 5), a few symbols are treated as words during POS tagging. This leads to discrepancies in sentence length because \texttt{evalb} discards symbols in the gold parse tree during evaluation.
Our proposed solution involves not disregarding any problematic labels and including symbols as words during evaluation. 
This approach implies that POS tagging results are based on the entire token numbers. It is noteworthy that \texttt{evalb}'s POS tagging results are rooted in the number of words, excluding symbols. 
The two remaining cases (sentences 3 and 4) involve actual word mismatches where trace symbols (*-\textit{num}) are inserted into the sentences. 
Naturally, \texttt{evalb} cannot handle these cases due to word mismatches. However, as we explained, our proposed algorithm addresses this issue by performing word alignment after sentence alignment.

\begin{figure}[!ht]
\centering
\resizebox{.49\textwidth}{!}{\footnotesize
\texttt{\begin{tabular}{rrr rrr rrr rrr}
\multicolumn{12}{c}{}\\
\multicolumn{2}{c}{Sent}& &&& Mt &  \multicolumn{2}{c}{Br} & Cr&& Co& Tag\\
ID&  L&  St& Re&  Pr&  Br& gd& te& Br& Wd&  Tg&  Acc \\
\hdashline
1&   37&    0&   77.27&  62.96&    17&     22&   27&      5&     37&    30 &   81.08 \\
2&   21&    0&   69.23&  60.00&     9&     13&   15&      2&     21&    17&    80.95 \\
   3&   47&    0&   77.78&  80.00&    28&     36&   35&      4&     48&    43&    89.58\\
   4&   26&    0&   33.33&  35.29&     6   &  18&   17&      8&     27&    19&    70.37\\
   5&   44&    0&   42.31&  32.35&    11&     26&   34&     17&     44&    33&    75.00 \\
\end{tabular}
}}
\caption{Evaluation results of bug cases by \texttt{evalb}}
\label{evalb-bug}
\end{figure}

\paragraph{Korean end-to-end parsing evaluation}
We conduct a comprehensive parsing evaluation for Korean, using system-segmented sequences as input for constituency parsing. These sequences may deviate from the corresponding gold standard sentences and tokens. 
We utilized the following resources for our parsing evaluation to simulate the end-to-end process:
(i) A set of 148 test sentences with 4538 tokens (morphemes)  from \texttt{BGAA0001} of the Korean Sejong treebank, as detailed in \citet{kim-park:2022}. In the present experiment, all sentences have been merged into a single text block.
(ii) POS tagging performed by \texttt{sjmorph.model} \citep{park-tyers:2019:LAW} for morpheme segmentation.\footnote{\url{https://github.com/jungyeul/sjmorph}}
The model's pipeline includes sentence boundary detection and tokenization through morphological analysis, generating an input format for the parser. 
(iii) A Berkeley parser model for Korean trained on the Korean Sejong treebank \citep{park-hong-cha:2016:PACLIC}.\footnote{\url{https://zenodo.org/records/3995084}}.
Figure~\ref{korean-evalb} presents the showcase results of end-to-end Korean constituency parsing. 
Given our sentence boundary detection and tokenization processes, there is a possibility of encountering sentence and word mismatches during constituency parsing evaluation. 
The system results show 123 sentences and 4367 morphemes because differences in sentence boundaries and tokenization results. 
During the evaluation, \texttt{jp-evalb} successfully aligns even in the presence of sentence and word mismatches, and subsequently, the results of constituency parsing are assessed.

\begin{figure}[!ht]
\centering
\resizebox{.49\textwidth}{!}{\footnotesize
\texttt{\begin{tabular}{rrr rrr rrr rrr}
\multicolumn{12}{c}{}\\
\multicolumn{2}{c}{Sent}& &&& Mt &  \multicolumn{2}{c}{Br} & Cr&& Co& Tag\\
ID&  L&  St& Re&  Pr&  Br& gd& te& Br& Wd&  Tg&  Acc \\
\hdashline
1&28& 0&85.71 & 85.71& 18&  21&21&3&  29& 26& 89.66\\
2&27& 0&91.30 & 84.00& 21&  23&25&2&  28& 25& 89.29\\
3&33& 0&88.00 & 88.00& 22&  25&25&3&  35& 31& 88.57\\
4&43& 0&72.73 & 72.73& 24&  33&33&7&  43& 40& 93.02\\
5&18& 0&69.57 & 84.21& 16&  23&19&2&  19& 12& 63.16\\
\multicolumn{12}{l}{.....}
\end{tabular}
}}
\caption{Evaluation results of the end-to-end Korean constituency parsing}
\label{korean-evalb}
\end{figure}

\section{Previous Work}

\texttt{tedeval} \citep{tsarfaty-nivre-andersson:2012:ACL2012short} is built upon the tree edit distance (\textsc{add} and \textsc{del}) by \citet{bille-2005}, incorporating the numbers of nonterminal nodes in the system and gold trees.
\texttt{conllu\_eval}\footnote{\url{https://universaldependencies.org/conll18/conll18_ud_eval.py}} treats tokens and sentences as spans. In case of a mismatch in the span positions between the system and gold files on a character level, the file with a smaller start value will skip to the next token until there is no start value mismatch. Similar processes are applied to evaluating sentence boundaries.
For \texttt{sparseval} \citep{roark-etal-2006-sparseval}, a head percolation table \citep{collins:1999} identifies head-child relations between terminal nodes and calculates the dependency score. Unfortunately, \texttt{sparseval} is currently unavailable.
\texttt{evalb}, the constituency parsing evaluation metric for nearly thirty years, despite inherent problems, has been widely used. 


\section{Conclusion}

Despite the widespread use and acceptance of the previous PARSEVAL measure as the standard tool for constituency parsing evaluation, it faces a significant limitation by requiring specific task-oriented environments. Consequently, there is still room for a more robust and reliable evaluation approach. Various metrics have attempted to address issues related to word and sentence mismatches by employing complex tree operations or adopting dependency scoring methods. In contrast, our proposed method aligns sentences and words as a preprocessing step without altering the original PARSEVAL measures. This approach allows us to preserve the complexity of the original \texttt{evalb} implementation of PARSEVAL while introducing a linear time alignment process. Given the high compatibility of our method with existing PARSEVAL measures, it also ensures the consistency and seamless integration of previous work evaluated using PARSEVAL into our approach. Ultimately, this new measurement approach offers the opportunity to evaluate constituency parsing within an end-to-end pipeline, addressing discrepancies that may arise during earlier steps, such as tokenization and sentence boundary detection. This enables a more comprehensive evaluation of constituency parsing. All codes and results from the case studies can be accessed at \url{https://github.com/jungyeul/alignment-based-PARSEVAL/}.

\section*{Acknowledgement}
This research is based upon work partially supported by \textit{Students as Partners} for Eunkyul Leah Jo, and \textit{The Work Learn Program} for Angela Yoonseo Park at The University of British Columbia.


\begin{thebibliography}{33}
\expandafter\ifx\csname natexlab\endcsname\relax\def\natexlab#1{#1}\fi

\bibitem[{Bille(2005)}]{bille-2005}
Philip Bille. 2005.
\newblock \href {https://doi.org/https://doi.org/10.1016/j.tcs.2004.12.030} {{A
  survey on tree edit distance and related problems}}.
\newblock \emph{Theoretical Computer Science}, 337(1):217--239.

\bibitem[{Black et~al.(1991)Black, Abney, Flickinger, Gdaniec, Grishman,
  Harrison, Hindle, Ingria, Jelinek, Klavans, Liberman, Marcus, Roukos,
  Santorini, and Strzalkowski}]{black-etal-1991-procedure}
Ezra Black, Steve Abney, Dan Flickinger, Claudia Gdaniec, Ralph Grishman, Phil
  Harrison, Donald Hindle, Robert Ingria, Frederick Jelinek, Judith~L. Klavans,
  Mark Liberman, Mitch Marcus, Salim Roukos, Beatrice Santorini, and Tomek
  Strzalkowski. 1991.
\newblock \href {https://aclanthology.org/H91-1060} {{A Procedure for
  Quantitatively Comparing the Syntactic Coverage of English Grammars}}.
\newblock In \emph{Speech and Natural Language: Proceedings of a Workshop Held
  at Pacific Grove, California, February 19-22, 1991}, pages 306--311, Pacific
  Grove, California. DARPA/ISTO.

\bibitem[{Brown et~al.(1993)Brown, Della~Pietra, Della~Pietra, and
  Mercer}]{brown-EtAl:1993:CL}
Peter~F. Brown, Stephen~A. Della~Pietra, Vincent~J. Della~Pietra, and Robert~L.
  Mercer. 1993.
\newblock \href {https://aclanthology.org/J93-2003} {{The Mathematics of
  Statistical Machine Translation: Parameter Estimation}}.
\newblock \emph{Computational Linguistics}, 19(2):263--311.

\bibitem[{Collins(1999)}]{collins:1999}
Michael Collins. 1999.
\newblock \href {http://www.cs.columbia.edu/~mcollins/papers/thesis.ps}
  {\emph{{Head-Driven Statistical Models for Natural Language Parsing}}}.
\newblock Ph.D. thesis, University of Pennsylvania.

\bibitem[{DeNero and Klein(2007)}]{denero-klein-2007-tailoring}
John DeNero and Dan Klein. 2007.
\newblock \href {https://aclanthology.org/P07-1003} {{Tailoring Word Alignments
  to Syntactic Machine Translation}}.
\newblock In \emph{Proceedings of the 45th Annual Meeting of the Association of
  Computational Linguistics}, pages 17--24, Prague, Czech Republic. Association
  for Computational Linguistics.

\bibitem[{Fern{\'{a}}ndez-Gonz{\'{a}}lez and
  G{\'{o}}mez-Rodr{\'{i}}guez(2020)}]{fernandez-gonzalez-gomez-rodriguez-2020-enriched}
Daniel Fern{\'{a}}ndez-Gonz{\'{a}}lez and Carlos G{\'{o}}mez-Rodr{\'{i}}guez.
  2020.
\newblock \href {https://doi.org/10.18653/v1/2020.acl-main.376} {{Enriched
  In-Order Linearization for Faster Sequence-to-Sequence Constituent Parsing}}.
\newblock In \emph{Proceedings of the 58th Annual Meeting of the Association
  for Computational Linguistics}, pages 4092--4099, Online. Association for
  Computational Linguistics.

\bibitem[{Gale and Church(1993)}]{gale-church:1993:CL}
William~A. Gale and Kenneth~W. Church. 1993.
\newblock \href {https://aclanthology.org/J93-1004} {{A Program for Aligning
  Sentences in Bilingual Corpora}}.
\newblock \emph{Computational Linguistics}, 19(1):75--102.

\bibitem[{Jo et~al.(2024)Jo, Park, and Park}]{jo-park-park-2024-evalb}
Eunkyul~Leah Jo, Angela~Yoonseo Park, and Jungyeul Park. 2024.
\newblock \href {https://doi.org/10.1162/coli{\_}a{\_}00512} {{A Novel
  Alignment-based Approach for PARSEVAL Measures}}.
\newblock \emph{Computational Linguistics}, pages 1--10.

\bibitem[{Kasai et~al.(2017)Kasai, Frank, McCoy, Rambow, and
  Nasr}]{kasai-etal-2017-tag}
Jungo Kasai, Bob Frank, Tom McCoy, Owen Rambow, and Alexis Nasr. 2017.
\newblock \href {https://doi.org/10.18653/v1/D17-1180} {{TAG Parsing with
  Neural Networks and Vector Representations of Supertags}}.
\newblock In \emph{Proceedings of the 2017 Conference on Empirical Methods in
  Natural Language Processing}, pages 1712--1722, Copenhagen, Denmark.
  Association for Computational Linguistics.

\bibitem[{Kasai et~al.(2018)Kasai, Frank, Xu, Merrill, and
  Rambow}]{kasai-etal-2018-end}
Jungo Kasai, Robert Frank, Pauli Xu, William Merrill, and Owen Rambow. 2018.
\newblock \href {https://doi.org/10.18653/v1/N18-1107} {{End-to-End Graph-Based
  TAG Parsing with Neural Networks}}.
\newblock In \emph{Proceedings of the 2018 Conference of the North American
  Chapter of the Association for Computational Linguistics: Human Language
  Technologies, Volume 1 (Long Papers)}, pages 1181--1194, New Orleans,
  Louisiana. Association for Computational Linguistics.

\bibitem[{Kim and Park(2022)}]{kim-park:2022}
Mija Kim and Jungyeul Park. 2022.
\newblock \href {https://doi.org/10.1017/S1351324920000479} {{A note on
  constituent parsing for Korean}}.
\newblock \emph{Natural Language Engineering}, 28(2):199--222.

\bibitem[{Lee et~al.(2016)Lee, Lewis, and
  Zettlemoyer}]{lee-lewis-zettlemoyer:2016:EMNLP}
Kenton Lee, Mike Lewis, and Luke Zettlemoyer. 2016.
\newblock \href {https://aclweb.org/anthology/D16-1262} {{Global Neural CCG
  Parsing with Optimality Guarantees}}.
\newblock In \emph{Proceedings of the 2016 Conference on Empirical Methods in
  Natural Language Processing}, pages 2366--2376, Austin, Texas. Association
  for Computational Linguistics.

\bibitem[{Lewis et~al.(2016)Lewis, Lee, and
  Zettlemoyer}]{lewis-lee-zettlemoyer:2016:NAACL}
Mike Lewis, Kenton Lee, and Luke Zettlemoyer. 2016.
\newblock \href {http://www.aclweb.org/anthology/N16-1026} {{LSTM CCG
  Parsing}}.
\newblock In \emph{Proceedings of the 2016 Conference of the North American
  Chapter of the Association for Computational Linguistics: Human Language
  Technologies}, pages 221--231, San Diego, California. Association for
  Computational Linguistics.

\bibitem[{Liang et~al.(2006)Liang, Taskar, and
  Klein}]{liang-etal-2006-alignment}
Percy Liang, Ben Taskar, and Dan Klein. 2006.
\newblock \href {https://aclanthology.org/N06-1014} {{Alignment by Agreement}}.
\newblock In \emph{Proceedings of the Human Language Technology Conference of
  the {\{}NAACL{\}}, Main Conference}, pages 104--111, New York City, USA.
  Association for Computational Linguistics.

\bibitem[{Liu and Zhang(2017{\natexlab{a}})}]{liu-zhang-2017-encoder}
Jiangming Liu and Yue Zhang. 2017{\natexlab{a}}.
\newblock \href {https://aclanthology.org/W17-6315} {{Encoder-Decoder
  Shift-Reduce Syntactic Parsing}}.
\newblock In \emph{Proceedings of the 15th International Conference on Parsing
  Technologies}, pages 105--114, Pisa, Italy. Association for Computational
  Linguistics.

\bibitem[{Liu and Zhang(2017{\natexlab{b}})}]{liu-zhang-2017-shift}
Jiangming Liu and Yue Zhang. 2017{\natexlab{b}}.
\newblock \href {https://transacl.org/ojs/index.php/tacl/article/view/927}
  {{Shift-Reduce Constituent Parsing with Neural Lookahead Features}}.
\newblock \emph{Transactions of the Association for Computational Linguistics},
  5:45--58.

\bibitem[{Moore(2002)}]{moore:2002}
Robert~C. Moore. 2002.
\newblock {Fast and Accurate Sentence Alignment of Bilingual Corpora}.
\newblock In \emph{Proceedings of the 5th Conference of the Association for
  Machine Translation in the Americas on Machine Translation: From Research to
  Real Users}, pages 135--244, Tiburon, CA, USA. Springer-Verlag.

\bibitem[{Och and Ney(2000)}]{och-ney:2000:ACL}
Franz~Josef Och and Hermann Ney. 2000.
\newblock \href {http://www.aclweb.org/anthology/N/N09/N09-1069} {{Improved
  Statistical Alignment Models}}.
\newblock In \emph{Proceedings of the 38th Annual Meeting of the Association
  for Computational Linguistics}, pages 440--447, Hong Kong. Association for
  Computational Linguistics.

\bibitem[{Och and Ney(2003)}]{och-ney:2003:CL}
Franz~Josef Och and Hermann Ney. 2003.
\newblock {A Systematic Comparison of Various Statistical Alignment Models}.
\newblock \emph{Computational Linguistics}, 29(1):19--51.

\bibitem[{Park et~al.(2016)Park, Hong, and Cha}]{park-hong-cha:2016:PACLIC}
Jungyeul Park, Jeen-Pyo Hong, and Jeong-Won Cha. 2016.
\newblock \href {http://aclweb.org/anthology/Y/Y16/Y16-2002.pdf} {{Korean
  Language Resources for Everyone}}.
\newblock In \emph{Proceedings of the 30th Pacific Asia Conference on Language,
  Information and Computation: Oral Papers (PACLIC 30)}, pages 49--58, Seoul,
  Korea. Pacific Asia Conference on Language, Information and Computation.

\bibitem[{Park and Tyers(2019)}]{park-tyers:2019:LAW}
Jungyeul Park and Francis Tyers. 2019.
\newblock \href {https://www.aclweb.org/anthology/W19-4022} {{A New Annotation
  Scheme for the Sejong Part-of-speech Tagged Corpus}}.
\newblock In \emph{Proceedings of the 13th Linguistic Annotation Workshop},
  pages 195--202, Florence, Italy. Association for Computational Linguistics.

\bibitem[{Petrov and Klein(2007)}]{petrov-klein:2007:main}
Slav Petrov and Dan Klein. 2007.
\newblock \href {http://www.aclweb.org/anthology/N/N07/N07-1051} {{Improved
  Inference for Unlexicalized Parsing}}.
\newblock In \emph{Human Language Technologies 2007: The Conference of the
  North American Chapter of the Association for Computational Linguistics;
  Proceedings of the Main Conference}, pages 404--411, Rochester, New York.
  Association for Computational Linguistics.

\bibitem[{Roark et~al.(2006)Roark, Harper, Charniak, Dorr, Johnson, Kahn, Liu,
  Ostendorf, Hale, Krasnyanskaya, Lease, Shafran, Snover, Stewart, and
  Yung}]{roark-etal-2006-sparseval}
Brian Roark, Mary Harper, Eugene Charniak, Bonnie Dorr, Mark Johnson, Jeremy
  Kahn, Yang Liu, Mari Ostendorf, John Hale, Anna Krasnyanskaya, Matthew Lease,
  Izhak Shafran, Matthew Snover, Robin Stewart, and Lisa Yung. 2006.
\newblock \href {http://www.lrec-conf.org/proceedings/lrec2006/pdf/116_pdf.pdf}
  {{SParseval: Evaluation Metrics for Parsing Speech}}.
\newblock In \emph{Proceedings of the Fifth International Conference on
  Language Resources and Evaluation (LREC'06)}, pages 333--338, Genoa, Italy.
  European Language Resources Association (ELRA).

\bibitem[{Seddah et~al.(2014)Seddah, K{\"{u}}bler, and
  Tsarfaty}]{seddah-kubler-tsarfaty:2014}
Djamé Seddah, Sandra K{\"{u}}bler, and Reut Tsarfaty. 2014.
\newblock \href {https://www.aclweb.org/anthology/W14-6111} {{Introducing the
  SPMRL 2014 Shared Task on Parsing Morphologically-rich Languages}}.
\newblock In \emph{Proceedings of the First Joint Workshop on Statistical
  Parsing of Morphologically Rich Languages and Syntactic Analysis of
  Non-Canonical Languages}, pages 103--109, Dublin, Ireland. Dublin City
  University.

\bibitem[{Seddah et~al.(2013)Seddah, Tsarfaty, K{\"{u}}bler, Candito, Choi,
  Farkas, Foster, Goenaga, Gojenola~Galletebeitia, Goldberg, Green, Habash,
  Kuhlmann, Maier, Nivre, Przepi{\'{o}}rkowski, Roth, Seeker, Versley, Vincze,
  Woli{\'{n}}ski, Wr{\'{o}}blewska, and de~la
  Clergerie}]{seddah-EtAl:2013:SPMRL}
Djamé Seddah, Reut Tsarfaty, Sandra K{\"{u}}bler, Marie Candito, Jinho~D.
  Choi, Richárd Farkas, Jennifer Foster, Iakes Goenaga, Koldo
  Gojenola~Galletebeitia, Yoav Goldberg, Spence Green, Nizar Habash, Marco
  Kuhlmann, Wolfgang Maier, Joakim Nivre, Adam Przepi{\'{o}}rkowski, Ryan Roth,
  Wolfgang Seeker, Yannick Versley, Veronika Vincze, Marcin Woli{\'{n}}ski,
  Alina Wr{\'{o}}blewska, and Eric~Villemonte de~la Clergerie. 2013.
\newblock \href {http://www.aclweb.org/anthology/W13-4917} {{Overview of the
  SPMRL 2013 Shared Task: A Cross-Framework Evaluation of Parsing
  Morphologically Rich Languages}}.
\newblock In \emph{Proceedings of the Fourth Workshop on Statistical Parsing of
  Morphologically-Rich Languages}, pages 146--182, Seattle, Washington, USA.
  Association for Computational Linguistics.

\bibitem[{Sennrich and Volk(2011)}]{sennrich-volk-2011-iterative}
Rico Sennrich and Martin Volk. 2011.
\newblock \href {https://aclanthology.org/W11-4624} {{Iterative, MT-based
  Sentence Alignment of Parallel Texts}}.
\newblock In \emph{Proceedings of the 18th Nordic Conference of Computational
  Linguistics (NODALIDA 2011)}, pages 175--182, Riga, Latvia. Northern European
  Association for Language Technology (NEALT).

\bibitem[{Stanojevi{\'{c}} and Steedman(2020)}]{stanojevic-steedman:2020:ACL}
Miloš Stanojevi{\'{c}} and Mark Steedman. 2020.
\newblock \href {https://doi.org/10.18653/v1/2020.acl-main.378} {{Max-Margin
  Incremental CCG Parsing}}.
\newblock In \emph{Proceedings of the 58th Annual Meeting of the Association
  for Computational Linguistics}, pages 4111--4122, Online. Association for
  Computational Linguistics.

\bibitem[{Thompson and Koehn(2019)}]{thompson-koehn-2019-vecalign}
Brian Thompson and Philipp Koehn. 2019.
\newblock \href {https://doi.org/10.18653/v1/D19-1136} {{Vecalign: Improved
  Sentence Alignment in Linear Time and Space}}.
\newblock In \emph{Proceedings of the 2019 Conference on Empirical Methods in
  Natural Language Processing and the 9th International Joint Conference on
  Natural Language Processing (EMNLP-IJCNLP)}, pages 1342--1348, Hong Kong,
  China. Association for Computational Linguistics.

\bibitem[{Tsarfaty et~al.(2012)Tsarfaty, Nivre, and
  Andersson}]{tsarfaty-nivre-andersson:2012:ACL2012short}
Reut Tsarfaty, Joakim Nivre, and Evelina Andersson. 2012.
\newblock \href {http://www.aclweb.org/anthology/P12-2002} {{Joint Evaluation
  of Morphological Segmentation and Syntactic Parsing}}.
\newblock In \emph{Proceedings of the 50th Annual Meeting of the Association
  for Computational Linguistics (Volume 2: Short Papers)}, pages 6--10, Jeju
  Island, Korea. Association for Computational Linguistics.

\bibitem[{Varga et~al.(2005)Varga, N{\'{e}}meth, Hal{\'{a}}csy, Kornai,
  Tr{\'{o}}n, and Nagy}]{varga-EtAl:2005}
Dániel Varga, Lázló N{\'{e}}meth, Péter Hal{\'{a}}csy, András Kornai,
  Viktor Tr{\'{o}}n, and Viktor Nagy. 2005.
\newblock {Parallel corpora for medium density languages}.
\newblock In \emph{Proceedings of the RANLP (Recent Advances in Natural
  Language Processing)}, pages 590--596, Borovets, Bulgaria.

\bibitem[{Vinyals et~al.(2015)Vinyals, Kaiser, Koo, Petrov, Sutskever, and
  Hinton}]{vinyals-EtAl:2015:NIPS}
Oriol Vinyals, Lukasz Kaiser, Terry Koo, Slav Petrov, Ilya Sutskever, and
  Geoffrey~E. Hinton. 2015.
\newblock \href
  {http://papers.nips.cc/paper/5635-grammar-as-a-foreign-language.pdf}
  {{Grammar as a Foreign Language}}.
\newblock In C.~Cortes, N.~D. Lawrence, D.~D. Lee, M.~Sugiyama, and R.~Garnett,
  editors, \emph{Advances in Neural Information Processing Systems 28}, pages
  2773--2781. Curran Associates, Inc.

\bibitem[{Wei et~al.(2020)Wei, Wu, and Lan}]{wei-etal-2020-span}
Yang Wei, Yuanbin Wu, and Man Lan. 2020.
\newblock \href {https://doi.org/10.18653/v1/2020.acl-main.299} {{A Span-based
  Linearization for Constituent Trees}}.
\newblock In \emph{Proceedings of the 58th Annual Meeting of the Association
  for Computational Linguistics}, pages 3267--3277, Online. Association for
  Computational Linguistics.

\bibitem[{Yamaki et~al.(2023)Yamaki, Taniguchi, and
  Mochihashi}]{yamaki-etal-2023-holographic}
Ryosuke Yamaki, Tadahiro Taniguchi, and Daichi Mochihashi. 2023.
\newblock \href {https://doi.org/10.18653/v1/2023.acl-long.15} {{Holographic
  CCG Parsing}}.
\newblock In \emph{Proceedings of the 61st Annual Meeting of the Association
  for Computational Linguistics (Volume 1: Long Papers)}, pages 262--276,
  Toronto, Canada. Association for Computational Linguistics.

\end{thebibliography}

\end{document}